# On the Geometry of Bayesian Graphical Models with Hidden Variables

**Raffaella Settimi**
Department of Statistics,
University of Warwick,
Coventry CV4 7AL (U.K.).
e-mail: r.settimi@warwick.ac.uk

**Jim Q. Smith**
Department of Statistics,
University of Warwick,
Coventry CV4 7AL (U.K.).
e-mail: j.q.smith@warwick.ac.uk

## Abstract

In this paper we investigate the geometry of the likelihood of the unknown parameters in a simple class of Bayesian directed graphs with hidden variables. This enables us, before any numerical algorithms are employed, to obtain certain insights in the nature of the unidentifiability inherent in such models, the way posterior densities will be sensitive to prior densities and the typical geometrical form these posterior densities might take. Many of these insights carry over into more complicated Bayesian networks with systematic missing data.

## 1 Introduction

The problem of learning about Bayesian networks from a data set is of great interest. Both the selection of an appropriate Bayesian network and the estimation of the probabilities that parametrise such graphical models are typically more complicated when some variables are hidden and have given rise to growing attention. For instance the problem of model selection for such a class of models is studied by Geiger et al. (1996) that propose an approximate Bayesian information criterion which depends on the dimension of the model.

When the sample is from a multinomial distribution, a common prior assumption for Bayesian learning is that the model parameters have a Dirichlet distribution. When the parameter estimation is based on complete data sets or data on ancestor sets, posterior distributions can then be calculated in closed form (see e.g. Spiegelhalter and Lauritzen 1990, Geiger and Heckerman 1997). However when some data is missing, the Bayesian probabilistic updating yields a posterior distribution which is a discrete mixture of Dirichlet distributions. In general the number of terms in this mixture may explode and dependencies are introduced a posteriori across the individual components of the vector of conditional probabilities. Consequently the computation of the posterior distribution often becomes intractable. Several approximation techniques have been developed for handling such mixtures (Spiegelhalter and Cowell 1992, Ramoni and Sebastiani 1997, Cowell 1998). Furthermore numerical algorithms have been proposed by applying Markov Chain Monte Carlo methods (see e.g. Neal 1993), or using the EM algorithm (Lauritzen 1995). Such methods appear to be promising when data is missing at random.

In practice however there are often systematic dependencies in the censoring mechanism and the missing-at-random assumption is likely to be unrealistic. The application of numerical or approximating analytical learning algorithms when hidden variables occur can become extremely inefficient (Cowell 1998). Inference about the unobservable nodes in the graph is usually very sensitive to the precise form of the prior distribution over model parameters and may yield multiple probability estimates of the latent parameters which are very different and explain the data equally well.

To illustrate the effect of unobserved variables on Bayesian learning, we concentrate our attention to a simple class of discrete Bayesian graphical models, displayed in Figure 1, which have a single hidden variable $Y_2$ and two observed nodes $Y_1, Y_3$. The study of the geometry of the parameter space will allow us to gain a good understanding to which features in the prior the ensuing inference is sensitive as well as enables us to predict how bad the shape of the full posterior might become.

Some interesting classes of chain graphs and Bayes networks on discrete data conform this structure if we allow $Y_1, Y_2, Y_3$ to be vectors. The naive-Bayes model in 3 variables with hidden root is Markov equivalent to the class of models here considered.

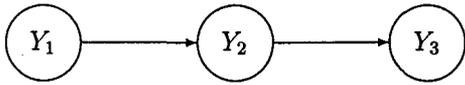

Figure 1: Directed graph with hidden variable $Y_2$

This conditional independence structure is applied also in latent structure analysis (Goodman 1974), used commonly in psychological and social models, where the observed variables are assumed statistically dependent but conditionally independent with respect to some latent variable. An important contribution to this field is given by Gilula (1979).

A clinical application of the conditional independence model of Figure 1 may arise when only data on the variable $Y_1$ representing patient's medical records (e.g. sex, age) and disease and on symptoms $Y_3$ are available. The direct physical consequences of diseases which may influence the effect of subsequent treatment, say $Y_2$, are not observed. One of the many instances of this type of model and data structure is given in the graphical model of Spiegelhalter and Cowell (1992) to diagnose congenital heart disease in newborn babies. Notice that unlike the latent variable models where the interest is focused only on the margins $(Y_1, Y_3)$, in application like this it may be essential for future diagnoses to draw inferences on $(Y_1, Y_2)$ and $(Y_2, Y_3)$ that are not directly observed from the data.

## 2 Geometry of the parameter space

Let $Y_1, Y_2, Y_3$ denote three categorical random variables. For notational convenience the levels of each variable are coded as positive integers so that $Y_i$ takes possible values $y_i = 1, ..., r_i$, for $i = 1, 2, 3$.

The multinomial model associated to $(Y_1, Y_2, Y_3)$ is an exponential model whose parameter space $\Theta$ is the simplex defined in $\mathbb{R}^{(r_1 r_2 r_3)}$ by

$$\Theta = \{\theta(i,j,k) > 0 \; 1 \leq i \leq r_1, \; 1 \leq j \leq r_2,$$
$$1 \leq k \leq r_3, \sum_{i,j,k} \theta(i,j,k) = 1\}$$

with dimension $d = r_1 r_2 r_3 - 1$, where the parameter $\theta(i,j,k)$ is the cell probability $P(Y_1 = i, Y_2 = j, Y_3 = k)$.

The conditional independence assumption $Y_1 \perp\!\!\!\perp Y_3 | Y_2$ induces constraints on the parameter space and the corresponding graphical model depicted in Figure 1 is a curved exponential model with dimension $t = r_1 r_2 + r_2 r_3 - r_2 - 1$. The dimension of an exponential model
Geometry of Bayesian Graphical Models 473

is the dimension of the parameter space considering its minimal parametrisation. Definitions and some results on multidimensional exponential models are recalled in the appendix.

The conditional independence assumption $Y_1 \perp\!\!\!\perp Y_3 | Y_2$ determines a set of non-linear constraints on the parameter space $\Theta$ since the probability model must satisfy

$$p(Y_1, Y_2, Y_3) p(Y_2) = p(Y_1, Y_2) p(Y_2, Y_3). \quad (1)$$

In $\Theta$ the probabilistic relationship (1) is equivalent to a system of $s = r_2(r_1 - 1)(r_3 - 1)$ irredundant quadratic equations expressed by

$$\theta(I, j, K)\theta(i, j, k) - \theta(I, j, k)\theta(i, j, K) = 0 \quad (2)$$

for $I, K$ denoting a fixed state $(Y_1 = I, Y_3 = K)$ for $i \neq I$, $1 \leq i \leq r_1$, $1 \leq j \leq r_2$ and $k \neq K$, $1 \leq k \leq r_3$.

The system of $s$ equations (2) defines a high dimensional algebraic variety $Q_s$ in $\Theta$ where the indeterminates $\theta(i,j,k)$ for $i = 1, ..., r_1, j = 1, ..., r_2, k = 1, ..., r_3$ are restricted to lie on. Thus the parameters space $\Theta_s \subset \Theta$ is defined by the intersection space $Q_s$ of the quadrics represented in (2).

The dimension of $Q_s$ is then $t = d - s = r_2(r_1 - 1) + r_2(r_3 - 1) + (r_2 - 1)$, i.e. $t$ is the dimension of the graphical model implying $Y_1 \perp\!\!\!\perp Y_3 | Y_2$.

We say that a parameter space $\Theta$ is identifiable by some data $y$ if for all $\theta, \theta'$ in $\Theta$, with $\theta \neq \theta'$, the probability distributions are such that $p(Y = y|\theta) \neq p(Y = y|\theta')$ for all $y$.

Our attention is focused on the particular situation where only a marginal two-way table on $(Y_1, Y_3)$ is observed. Thus the observed margins $(Y_1, Y_3)$ specify the linear space of identifiable parameters $\Theta_m$ in $\Theta_s$. The unidentifiable space, that is the space of parameters indistinguishable through the likelihood of the data, is then the complement space of $\Theta_m$ in $\Theta_s$, i.e. $\Theta_s \setminus \Theta_m$.

By using the singular value decomposition theorem Gilula (1979) shows that, if the dimension of the sample space satisfy $r_1, r_3 > 2$ and $r_2 = 2$, then there exist some marginal distributions $p(Y_1, Y_3)$ that cannot be parametrised through such latent variable models. He proposes a necessary and sufficient condition to determine whether a two-way table is consistent with the statement $Y_1 \perp\!\!\!\perp Y_3 | Y_2$ for $r_2 = 2$. The method requires the checking of two inequalities although the underlying geometrical implications of these are certainly not transparent.

It is well known that if $r_2 \geq \min(r_1, r_3)$ then the statement $Y_1 \perp\!\!\!\perp Y_3 | Y_2$ imposes no constraint on the table of probabilities $p(Y_1, Y_3)$. Thus, in particular, the dimension of the marginal space $\Theta_m$ is full, i.e. $m = r_1 r_3 - 1$. The converse of this result is also true.

The following example gives a conditional distribution $p(Y_3|Y_1)$ which is consistent with the conditional independence statement $Y_1 \perp\!\!\!\perp Y_3 | Y_2$ if $r_2 \geq \min(r_1, r_3)$.

**Example 1** Suppose the joint distribution $p(Y_1, Y_3)$ is such that $r_3 \geq r_1$ and $Y_3|Y_1$ is of the form

$$p(Y_3 = k | Y_1 = i) = \begin{cases} 1 & \text{if } 1 \leq i = k \leq r_1 \\ 0 & \text{if } 1 \leq i \neq k \leq r_3 \end{cases}$$

The conditional distribution $p(Y_3|Y_1)$ satisfies the relationship

$$p(Y_3 = k | Y_1 = i) = \sum_{j=1}^{r_2} p(Y_3 = k | Y_2 = j) p(Y_2 = j | Y_1 = i) \quad (3)$$

for $i = 1, ..., r_1, k = 1, ..., r_3$. Then setting $k = i$ in equation (3) and summing over $i$ we have

$$r_1 = \sum_{i=1}^{r_1} \sum_{j=1}^{r_2} p(Y_3 = i | Y_2 = j) p(Y_2 = j | Y_1 = i) \leq r_2.$$

That is the conditional distribution $p(Y_3|Y_1)$ is consistent with the conditional independence statement if and only if $r_2 \geq r_1$.

Therefore there must exist a value $j(i)$ of $Y_2$ such that

$$p(Y_3 = i | Y_2 = j(i)) = p(Y_2 = j(i) | Y_1 = i) = 1$$

so that

$$\begin{cases} p(Y_3 = i | Y_2 = j(i)) = 1 \\ p(Y_3 = k | Y_2 = j(i)) = 0 \quad 1 \leq k \leq r_3 \;\; k \neq i \end{cases}$$

$$\begin{cases} p(Y_2 = j(i) | Y_1 = i) = 1 \\ p(Y_2 = j | Y_1 = i) = 0 \quad 1 \leq j \leq r_2 \;\; j \neq j(i) \end{cases}$$

In particular, the required form of the conditional probability $p(Y_2|Y_1)$ demands that each value $j(i)$ of $Y_2$ can be associated with at most one value $i$ of $Y_1$.
□

This can be regarded as a positive result, because, by assuming our conditional independence statement, we implicitly impose constraints on what we can observe provided $r_2$ is small. However if data is consistent with our conditional independence model, then provided that the corresponding marginal probability table $p(Y_1, Y_3)$ is non-degenerate in a sense defined below, there will be a whole set of parameters which explain the observed data equally well. This is not so positive as these different parameter values will typically say different things about the underlying probability structure of the problem.

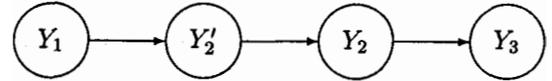

Figure 2: Directed graph with two hidden variables $Y_2, Y_2'$

**Definition 1** We say that a marginal distribution $p(Y_1, Y_3)$ is a regular point in $\Theta_m$ if it can be expressed as

$$p(Y_1 = i, Y_3 = k) = \\ p(Y_1 = i) \sum_{j=1}^{r_2} p(Y_3 = k | Y_2 = j) p(Y_2 = j | Y_1 = i)$$

for $1 \leq i \leq r_1, 1 \leq k \leq r_3$ where either $p(Y_3|Y_2)$ or $p(Y_2|Y_1)$ are non-degenerate, i.e. all the probability values are greater than zero.

Consider the conditional independence model displayed in Figure 2 where $Y_2'$ is a discrete latent random variable with $r_2$ states. Such a model implies that $Y_1 \perp\!\!\!\perp Y_3 | Y_2$ and also $Y_1 \perp\!\!\!\perp Y_3 | Y_2'$ so the pairs $p(Y_3|Y_2), p(Y_2|Y_1)$ and $p(Y_3|Y_2'), p(Y_2'|Y_1)$ will both be candidate for explaining the conditional distribution $p(Y_3|Y_1)$ and hence the marginal table $p(Y_1, Y_3)$.

**Example 2** Given the non-degenerate distributions $(p(Y_2|Y_1), p(Y_3|Y_2))$, we show how to construct a two-dimensional family of new latent variables $Y_2'$ for which the conditional probability distribution $p(Y_2|Y_2')$ is not degenerate. Construct $Y_2'$ so that the conditional distribution $p(Y_2|Y_2')$ is given by the transition matrix

$$Q = \begin{pmatrix} \pi & 1 - \pi \\ \rho & 1 - \rho \end{pmatrix} \quad \rho \neq \pi$$

so that for each $i, 1 \leq i \leq r_1$

$$(p(Y_2 = 1 | Y_1 = i), p(Y_2 = 2 | Y_1 = i)) = \\ (p(Y_2' = 1 | Y_1 = i), p(Y_2' = 2 | Y_1 = i)) Q$$

The distribution $p(Y_2'|Y_1)$ is obtained by inverting the relationship above and it can be expressed as

$$\begin{aligned} p(Y_2' = 1 | Y_1 = i) = & \; (\pi - \rho)^{-1} \times \\ & [(1 - \rho) p(Y_2 = 1 | Y_1 = i) - \\ & \rho p(Y_2 = 2 | Y_1 = i)] \\ p(Y_2' = 2 | Y_1 = i) = & \; (\pi - \rho)^{-1} \times \\ & [-(1 - \pi) p(Y_2 = 1 | Y_1 = i) + \\ & \pi p(Y_2 = 2 | Y_1 = i)]. \end{aligned}$$

for $i = 1, ..., r_1$

Without loss of generality set $\rho < \pi$. Thus $p(Y_2'|Y_1)$ is a distribution consistent with the marginal distribution $p(Y_1, Y_3)$ if and only if

$$0 \leq \rho \leq \min_{1 \leq i \leq r_1} p(Y_2 = 1 | Y_1 = i) \leq \\ \max_{1 \leq i \leq r_1} p(Y_2 = 1 | Y_1 = i) \leq \pi \leq 1$$

This clearly defines a two dimensional space of possible distributions $p(Y_2'|Y_1)$ and $p(Y_3|Y_2')$. Notice that by setting $\pi$ and $\rho$ on the boundary, i.e.

$$\rho = p(Y_2 = 1|Y_1 = i_1) = \min_{1 \leq i \leq r_1} p(Y_2 = 1|Y_1 = i)$$
$$\pi = p(Y_2 = 1|Y_1 = i_2) = \max_{1 \leq i \leq r_1} p(Y_2 = 1|Y_1 = i),$$

we obtain a candidate pair $(p(Y_2'|Y_1), p(Y_3|Y_2'))$ where $Y_2'$ is most informative about $Y_1$ and $Y_1 \perp\!\!\!\perp Y_3|Y_2'$. The conditional distribution $p(Y_2'|Y_1)$ becomes degenerate with $p(Y_2' = 1|Y_1 = i_1) = 0$ or $p(Y_2' = 2|Y_1 = i_2) = 0$ This result means that the margin $p(Y_1, Y_2')$ will have a zero in both columns of $p(Y_2')$. Symmetrical arguments permit to construct a random variable $Y_2^*$ most informative about $Y_3$ such that $Y_1 \perp\!\!\!\perp Y_3|Y_2^*$ where the margin $p(Y_2^*, Y_3)$ has a zero in each row of $p(Y_2^*)$. Note that a different family of solutions is obtained in an analogous way by demanding that $\pi < \rho$. □

Analogous result exists for all $r_1, r_2, r_3$. Thus

**Theorem 1** *For $r_2 < \min(r_1, r_3)$, if a regular point in $p(Y_1, Y_3)$ is observed the unidentifiable space in $\Theta_s$ has dimension $r_2(r_2 - 1)$.*

**Proof** The result follows from a straightforward generalisation of the arguments presented in Example 2. See Settimi and Smith (1998) for details.

It can also be shown (Settimi and Smith 1998) that, given a regular marginal distribution $p(Y_1, Y_3)$ parametrised in terms of the non degenerate distributions $(p(Y_2|Y_1), p(Y_3|Y_2))$, we can always find $Y_2'$ so that $p(Y_2'|Y_1)$ and hence $p(Y_1, Y_2')$ is degenerate. Analogously there exists a variable $Y_2^*$ such that $p(Y_2^*, Y_3)$ is degenerate.

Thus in all cases there is a probability parameter vector which has joint mass function over $(Y_1, Y_2, Y_3)$ with zeros in it which is at least as likely as any other possible probability distribution explaining the observed margins $(Y_1, Y_3)$.

## 3 A useful reparametrisation

We now consider an invertible transformation of the parameter space $\Theta$ into $\Delta \times \Lambda$ defined as

$$\theta(i, j, k) = \delta(i, k)\lambda_j(i, k) \quad (4)$$

for any $\theta(i, j, k) \in \Theta$ where the new parameters $\delta$'s and $\lambda$'s are the marginal and conditional probabilities:

$$\delta(i, k) = p(Y_1 = i, Y_3 = k),$$
$$\lambda_j(i, k) = p(Y_2 = j|Y_1 = i, Y_3 = k)$$

varying respectively in the simplex $\Delta$ of dimension $m = r_1 r_3 - 1$ and the set $\Lambda$ of dimension $l = r_1 r_3 (r_2 -$ 1). By applying (4) to equations (2), we can write

$$\delta(I, K)\delta(i, k)\lambda_j(I, K)\lambda_j(i, k) -$$
$$\delta(i, K)\delta(I, k)\lambda_j(i, K)\lambda_j(I, k) = 0 \quad (5)$$

for fixed states $I$ and $K$ and for $1 \leq i \neq I \leq r_1$ and $1 \leq k \neq K \leq r_3$. The restricted space $\Theta_s$, induced by the conditional independence assumption, is transformed into a subspace in $\Delta \times \Lambda$, specified by the algebraic variety defined by (5).

Suppose we observe the margins $(Y_1, Y_3)$, then the space $\Delta$ generated by the parameters $\delta(i, k)$ is identifiable.

*Case I:* $r_2 \geq \min(r_1, r_3)$

If $r_2 \geq \min(r_1, r_3)$ the conditional independence assumption does not affect the marginal space of $p(Y_1, Y_3)$ which has full dimension. A minimal parametrisation for $\theta_s$ can be defined by considering the parameters $\delta(i, k)$ in $\Delta$ and the $l - s$ parameters $\lambda_j(i, k)$, where $l - s$ is the codimension of the variety in $\Lambda$ defined by the equations (5).

The unidentifiable space has dimension $t - m = r_2(r_1 + r_3 - 1) - r_1 r_3$. This value is the dimension of the variety represented by the system of irredundant equations (5) in the space $\Theta_l$ of dimension $l = r_1 r_3 (r_2 - 1)$ generated by the indeterminates $\lambda$'s, where the dimension of the variety is calculated as the dimension of the embedding space less the number of the irredundant equations $l - s$.

The symmetric structure of the system (5) suggests to specify the solution space in terms of $l - s$ variables $\lambda_j(i, k)$ for some $i, j, k$, chosen so that equations (5) are linear in the selected indeterminates. Thus the unidentifiable space can be defined through a subset of $l - s$ indeterminates $\lambda$'s, bounds on these parameters must be specified as $\Theta_l$ is a probability space.

*Case II:* $r_2 < \min(r_1, r_3)$

The dimension of the unidentifiable space is $l = r_2(r_2 - 1)$ and the dimension of $\Theta_m$ is computed as $t - l = r_2(r_1 + r_3 - r_2) - 1$. Hence the conditional independence assumption imposes $(r_1 - r_2)(r_3 - r_2)$ constraints on the marginal space $p(Y_1, Y_3)$. Examples of such constraints are presented in section 4 for cases with $r_1 = r_3 = 3, r_2 = 2$ and $r_1 = r_3 = 4, r_2 = 3$

A full dimensional table $p(Y_1, Y_3)$ corresponds to sets of solutions of the system (5) that are contained on the boundary of $\Lambda$, i.e. families of degenerate conditional probabilities $p(Y_2|Y_1, Y_3)$. This set of solutions imposes some structural zeros in the joint probability table.

## 4 Some examples of unidentifiable spaces

In this section we use the reparametrisation described in Section 3 to investigate the geometry of the unidentifiable space in simple cases. The first example deals with three binary variables problem. Such a simple but not trivial case permits to show in detail how the probability space is constrained.

The second and third examples investigate the geometry of the parameter space for models where the unobserved variable has dimension smaller than the manifest variables.

### 4.1 Model with three binary variables

Let $Y_1, Y_2, Y_3$ be three binary variables. Suppose we have a table of counts on $(Y_1, Y_3)$, observations on $Y_2$ are completely missing.

The parameter space is restricted to the manifold expressed by

$$\begin{cases} \theta(1,1,1)\theta(2,1,2) - \theta(1,1,2)\theta(2,1,1) = 0 \\ \theta(1,2,1)\theta(2,2,2) - \theta(1,2,2)\theta(2,2,1) = 0 \end{cases} \quad (6)$$

Note that these two equations are the cross sums in the marginal two-way tables for $p(Y_1, Y_3|Y_2)$

Thus this model has parameter space $\Theta_s$ of dimension $t = 5$. As $Y_1$ and $Y_3$ have been observed, the marginal probabilities for these two variables define a linear 3-dimensional subspace of $\Theta$. Thus the model has at least a 2-dimensional unidentifiable parameter space.

Explicitly the unobservable space corresponds to the unlearnt probabilities $p(Y_2|Y_1, Y_3)$.

Reparametrise $\Theta_s$ by using the parameter transformation defined in (4) so that from equations (6) we obtain the following system

$$[1 - \lambda(1,1) - \lambda(2,2)]z - [1 - \lambda(1,2) - \lambda(2,1)] = 0 \quad (7)$$
$$\lambda(1,1)\lambda(2,2)z - \lambda(1,2)\lambda(2,1) = 0 \quad (8)$$

for $z \in \mathbb{R}_+$ defined by the cross-ratio of the marginal table

$$z = [\delta(1,1)\delta(2,2)]/[\delta(1,2)\delta(2,1)].$$

For convenience we have omitted the subscript $j = 1$ in the notation $\lambda(i,k) = \lambda_1(i,k)$. Under the hypothesis that the margins are observed, $z$ is a known positive scalar.

Notice that if $Y_1 \perp\!\!\!\perp Y_3$, the constant $z$ is equal to one and the above equations reduce to $\lambda(1,1) = \lambda(2,1), \lambda(2,2) = \lambda(1,2)$ that is the algebraic representation of a degenerate quadric. It says that for

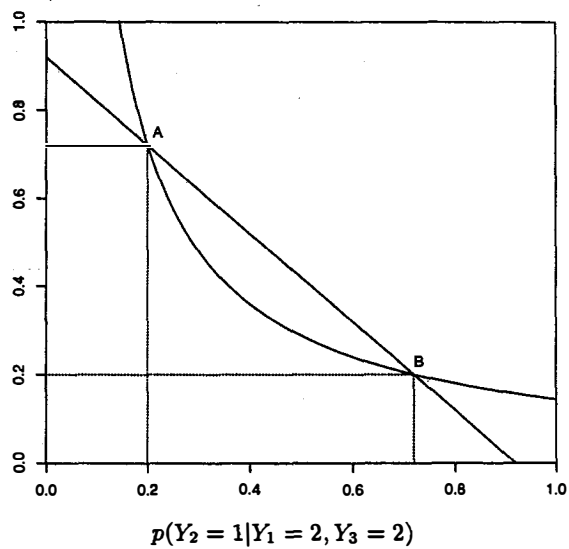

$p(Y_2 = 1|Y_1 = 1, Y_3 = 1)$

$p(Y_2 = 1|Y_1 = 2, Y_3 = 2)$

Figure 3: Plot of equations (10), (11) for $z = 0.8, c_1 = 0.3, c_2 = 0.6$. Intersection points are $A = (0.20, 0.72)$ and $B = (0.72, 0.20)$

$Y_1 \perp\!\!\!\perp Y_3$ the conditional probabilities $p(Y_2 = j|Y_1 = i, Y_3 = k)$ are all equal and therefore $p(Y_2 = j|Y_1 = i, Y_3 = k) = p(Y_2 = j)$.

The intersection of the hyperplane (7) and the non-degenerate quadric (8) determines a quadratic variety in the subspace $[0,1]^4$.

The relationships (7) and (8) are symmetric in $\lambda(1,1), \lambda(2,2)$ and $\lambda(1,2), \lambda(2,1)$ and simple algebra shows that the probabilities $p(Y_2 = 2|Y_1 = i, Y_3 = 1)$ for $i = 1, 2$ must lie on the surfaces

$$\begin{cases} \lambda(2,1) = \dfrac{[1 - \lambda(1,2) - z(1 - \lambda(2,2))]\lambda(2,2)}{\lambda(2,2) - \lambda(1,2)} \\ \lambda(1,1) = \dfrac{[1 - \lambda(1,2) - z(1 - \lambda(2,2))]\lambda(1,2)}{z(\lambda(2,2) - \lambda(1,2))} \end{cases}$$

with constraints on the variables $\lambda(2,2)$ and $\lambda(1,2)$ defined by

$$\begin{aligned} &\text{if } \lambda(2,2) > \lambda(1,2) \\ &\quad \frac{\lambda(1,2)}{\lambda(2,2)} < z < 1 \quad \frac{1 - \lambda(1,2)}{1 - \lambda(2,2)} > z > 1 \\ &\text{if } \lambda(2,2) < \lambda(1,2) \\ &\quad \frac{\lambda(1,2)}{\lambda(2,2)} > z > 1 \quad \frac{1 - \lambda(1,2)}{1 - \lambda(2,2)} < z < 1 \end{aligned} \quad (9)$$

Suppose $z > 1$, it is easy to check that, for any given probabilities $\lambda(2,1) = c_1$ and $\lambda(1,2) = c_2$ in $[0,1]$, $\lambda(1,1)$ and $\lambda(2,2)$ are not identifiable. The equations (7) and (8) become

$$\lambda(1,1) + z\lambda(2,2) - z - c_1 - c_2 + 1 = 0 \quad (10)$$
$$z\lambda(1,1)\lambda(2,2) - c_1 c_2 = 0 \quad (11)$$

whose algebraic solutions represent the intersection points $A(a_1, a_2)$ and $B(b_1, b_2)$ of the straight line (10) and the equilateral hyperbola (11) (see for example Fig. 3). The points $A$ and $B$ have symmetric coordinates, i.e. $a_1 = b_2, a_2 = b_1$. This indicates that the structure of the problem is invariant to the permutation of the events $\{Y_2 = 1\}$ and $\{Y_2 = 2\}$.

This will explain the aliasing in the estimates. The unidentifiability of the system is then of a rather "unpleasant" type. In fact if $p(Y_2 = 2|Y_1 = 2, Y_3 = 1)$ and $p(Y_2 = 2|Y_1 = 1, Y_3 = 2)$ were observed, then two symmetric equally likely estimates would be obtained for $p(Y_2 = 2|Y_1 = 1, Y_3 = 1)$ and $p(Y_2 = 2|Y_1 = 2, Y_3 = 2)$

Hence a minimal parametrisation for the unidentifiable space is determined by the variables $(\lambda(1,1), \lambda(1,2))$ or $(\lambda(2,1), \lambda(2,2))$ varying in $[0,1]$ with constraints expressed by (9).

### 4.2  Model for $r_1 = 3, r_2 = 2, r_3 = 3$

Suppose that the random variables $Y_1, Y_2, Y_3$ have states $r_1 = 3, r_2 = 2, r_3 = 3$ respectively. In this case the conditional independence statement $Y_1 \perp\!\!\!\perp Y_3|Y_2$ defines a 9-dimensional manifold $\Theta_s$ embedded in the 17-dimensional parameter space $\Theta$. This is the simplest case for $r_2 < \min(r_1, r_3)$. By fixing states $I = 1, K = 1$, the system of $s = 8$ equations (5) can be written as

$$\begin{cases} \delta(1,1)\delta(i,k)\lambda(1,1)\lambda(i,k) - \\ \delta(1,k)\delta(i,1)\lambda(1,k)\lambda(i,1) = 0 \\ \delta(1,1)\delta(i,k)(1 - \lambda(1,1) - \lambda(i,k)) - \\ \delta(1,k)\delta(i,1)(1 - \lambda(1,k)\lambda(i,1)) = 0 \end{cases} \quad (12)$$

for $i, k = 2, 3$.

The observed marginal probabilities $\delta$'s can be replaced in (12) with the cross-ratios defined by:

$$z_1 = \frac{\delta(1,1)\delta(2,2)}{\delta(1,2)\delta(2,1)}, \quad z_2 = \frac{\delta(1,1)\delta(2,3)}{\delta(1,3)\delta(2,1)},$$
$$z_3 = \frac{\delta(1,1)\delta(3,2)}{\delta(1,2)\delta(3,1)}, \quad z_4 = \frac{\delta(1,1)\delta(3,3)}{\delta(1,3)\delta(3,1)}$$

If the cross-ratios $z_1, ..., z_4$ are equal to one, that is $Y_1 \perp\!\!\!\perp Y_3$, it is readily seen from (12) that $\lambda(i,1) = \lambda(i,2) = \lambda(i,3)$ for $i = 1, 2, 3$ or $\lambda(1,k) = \lambda(2,k) = \lambda(3,k)$ for $k = 1, 2, 3$.

By solving the system of equations (12) with respect to the parameters $\lambda$'s and $z$'s, we find an algebraic relationship on the marginal probabilities given by

$$z_1 z_4 - z_2 z_3 = (z_1 + z_4) - (z_2 + z_3).$$

This constraint implies that the marginal space $\Theta_m$ is not full and that $\dim(\Theta_m) = 7$. The unidentifiable space $\Theta_l$ is 2-dimensional. The expression of the solution space is reported in the Appendix, where constraints on the two free indeterminates are also given.

The distributions $p(Y_1, Y_3)$ that do not satisfy the constraint above are parametrised by degenerate conditional distributions $(p(Y_1|Y_2), p(Y_2|Y_3))$. This corresponds to a solution space of (12) defined by degenerate distributions $\lambda(i,1) = \lambda(i,2) = \lambda(i,3) = 1$ for $i = 1, 2, 3$ or for symmetry $\lambda(i,1) = \lambda(i,2) = \lambda(i,3) = 0$ If we consider $\lambda(1,1) = 1, \lambda(1,2) = 1, \lambda(1,3) = 1$, the other components of the variety $Q_s$ are

$$\lambda(2,2) = \frac{1}{z_1}\lambda(2,1), \quad \lambda(2,3) = \frac{1}{z_2}\lambda(2,1),$$
$$\lambda(3,2) = \frac{1}{z_3}\lambda(3,1), \quad \lambda(3,3) = \frac{1}{z_4}\lambda(3,1)$$

Therefore each point in the unconstrained space of $p(Y_1, Y_3)$ is associated to a two-dimensional convex subset contained in the boundary of $\Lambda$.

Notice that such a particular degenerate distribution demands that some structural zeros are assigned to the joint distribution $p(Y_1, Y_2, Y_3)$, for instance $\lambda(1,1) = \lambda(1,2) = \lambda(1,3) = 1$ means that the joint probabilities $\theta(1,1,1) = \theta(1,1,2) = \theta(1,1,3) = 0$.

### 4.3  Model for $r_1 = 4, r_2 = 3, r_3 = 4$

In this example we study the parameter space of a Bayesian network where the observed variables have four states and the hidden variable has three states. The dimension of the unrestricted parameter space is 47. The conditional independence statement defines $s = 27$ quadratic equations in $\theta(y_1, y_2, y_3)$, thus the dimension of the manifold $Q_s$ associated to the 27 quadratic constraints is $t = 20$. The unidentifiable space is $r_2(r_2 - 1) = 6$ and the dimension of the marginal space is $m = 14$. The loss of one dimension of $\Theta_m$ is explained by the non-linear constraint found by solving equations (5). The solution space associated to unconstrained marginal spaces $p(Y_1, Y_3)$ is defined by degenerate distributions such as $\lambda_j(i,1) = \lambda_j(i,2) = \lambda_j(i,3) = 0$ for $j = 1, 2$ and fixed $i$. The same arguments discussed in the previous example for a binary latent variable hold also in this case.

## 5  Conclusions

One consequence of the results given above is that discrete Bayesian networks with hidden variables are very sensitive to the chosen form of prior densities over parameters. We have shown that typical likelihood functions will have flat ridges in them as well as several isolated maxima on the boundary of the parameter space of cell probabilities. The flat ridges will mean that the prior density behaviour over these regions will persist into the posterior and so will need to be set with great care. The isolated maxima on the boundaries will fight against the usual form of the families of prior densities like the composition of Dirichlet distributions (e.g. Geiger and Heckerman 1997). In particular it will make posterior distributions very sensitive to the tail behaviour of the prior, a feature which is not usually elicited with any degree of accuracy. This suggests us to be suspicious of the output of most routine Bayesian learning algorithms applied to networks of this type.

Aliasing problems, i.e. problems with multiple maxima, can largely be overcome by demanding order relations in the prior, but again such appropriate prior distributions would not be standard in form. For example there are no such Dirichlet priors. The problem of the bias towards degeneracy is very difficult to fix. A possible solution could be to set priors which demanded explanatory distributions lying on the boundary of the space. But this would require a complete reinterpretation of the family of graphical models considered.

Obviously these difficulties apply to more complicated Bayesian networks. For instance it can be shown that the $W$ graphical structure considered in Geiger et al. (1996) with a binary hidden variable is characterised by a 2-dimensional unidentifiable space independently on the number of states of the four observed variables. In a large system with hidden nodes it can be difficult to recognise the unidentifiability and aliasing problems that affect the Bayesian learning process leading to inefficient updating algorithms. However by studying the geometry induced by a Bayesian network we may be able to systematically identify how and when estimation problems are going to occur.

## Acknowledgements

This research was supported by the Engineering and Physical Sciences Research Council (Grant no. GR/K72254).

## Appendix

### A1 Overview on curved exponential models

Let $Y_1, Y_2, ..., Y_n$ denote $n$ categorical variables. The possible values $y_i$ taken by $Y_i$ are coded for convenience as integers such as $1 \leq y_i \leq r_i$ for $= 1, ..., n$. We assume that the random vector $(Y_1, ..., Y_n)$ has multinomial distribution with parameters $\theta(y_1, ...y_n) = p(Y_1 = y_1, ..., Y_n = y_n)$ associated to each state $(y_1, ..., y_n)$. The corresponding multinomial model $\mathcal{M}$ is exponential with parameter space

$$\Theta = \{\theta(y_1, ..., y_n) > 0 : \sum_{y_1, ..., y_n} \theta(y_1, ..., y_n) = 1\},$$

defined in the simplex in $\mathbb{R}^{d+1}$ of dimension $d = \prod_{i=1}^{n} r_i$.

Suppose now that the relationships among the variables can be described through a *directed acyclic graph* (DAG) $\mathcal{G}$ with nodes $Y_i$, $i = 1, ..., n$. Let $\Pi_i$ denote the set of parents of the variable $Y_i$ and assume that the variables are ordered so that there may be an edge from $Y_i$ pointing to $Y_j$ only if $i < j$.

Assuming a directed acyclic graph $\mathcal{G}$ for the variables $Y_1, ..., Y_n$ imposes non linear constraints on the parameter space $\Theta$ corresponding to the conditional independence assumptions implicit in the graph. The state space of $\mathcal{G}$ is embedded in $\Theta$ and can be parametrised through coordinates

$$\theta(y_i|\pi_i) = p(Y_i = y_i|\Pi_i = \pi_i) \text{ for } 1 \leq i \leq n \\ \text{with } \theta(y_i|\pi_i) > 0 \text{ and } \sum_{y_i=1}^{r_i} \theta(y_i|\pi_i) = 1 \quad (13)$$

where $\pi_i$ are the possible values of parents $\Pi_i$. Note that the number of parameters $\theta(y_i|\pi_i)$ is

$$t = \sum_{i=1}^{n} \dim(\Pi_i)[\dim(Y_i) - 1] \quad (14)$$

setting $\dim(\Pi_i) = 1$ if $\Pi_i = \emptyset$.

Theorem 2 states that the parameterisation (13) is minimal by showing that a DAG is a curved exponential model of dimension $t$. The definition of curved exponential model is recalled below (Kass and Vos, 1997, Ch. 4).

**Definition 2** *Let $S$ be a $k$-dimensional exponential family with parameter space $N$ such that $S = \{p_\eta : \eta \in N\}$. Assume there is a mapping $\theta \to \eta(\theta)$ for each $\theta \in \Theta_0$ that defines $N_0 = \eta(\Theta_0)$. The subfamily $S_0$ is a curved exponential family if $\Theta_0$ is an open subset in $\mathbb{R}^t$, and*

*(i) the mapping is one-to-one and smooth (infinitely differentiable), and of rank $t$, meaning that the $t \times k$ derivative matrix $D\eta(\theta)$ has rank $t$ everywhere;*

*(ii) writing $\rho : N_0 \to \Theta$ for the inverse mapping, if a sequence $\{\eta_n \in N_0\}$ converges to a point $\eta_0 \in N_0$, then the corresponding sequence $\{\rho(\eta_n) \in \Theta_0\}$ must converge to $\rho(\eta_0) \in \Theta_0$.*

**Theorem 2** *The DAG $\mathcal{G}$ is a curved exponential model with parameter space $\Theta_\mathcal{G}$ defined by the parametrisation in (13) whose dimension $t$ is calculated from (14).*

**Proof.** See Settimi and Smith (1998).

The following corollary extends the result of Theorem 2 to decomposable models.

**Corollary 1** *Let $\tilde{\mathcal{G}}$ denote an undirected decomposable graph with cliques $\{C_1, ..., C_m\}$ and separators $\{S_2, ..., S_m\}$, then the corresponding hierarchical model is a curved exponential family of dimension*

$$\sum_{i=1}^{m} \dim(C_i) - \sum_{i=2}^{m} \dim(S_i)$$

**Proof.** From Proposition 3.29 in Lauritzen (1996, p. 52) decomposable models are a subfamily of DAG models and hence from Theorem 2 decomposable models form a curved exponential family.

Furthermore in the decomposable models the cliques $C_i$ are the complete subsets $Y_j \cup \Pi_j$, hence from (14)

$$t = \sum_{i=1}^{n} \dim(\Pi_i)(\dim(Y_i)-1) = \sum_{i=1}^{n} \dim(C_i) - \sum_{i=2}^{n} \dim(S_i).$$

Note that the dimension of the decomposable models space is reported by Lauritzen (1996, p. 105) in Proposition 4.35.

### A2 Solution space for the model with $r_1 = r_3 = 3, r_2 = 2$

The solution of the system of equations (12) with respect to $\lambda(2,1)$ and $\lambda(2,2)$ is given by

$$\begin{cases} \lambda(1,1) = \dfrac{\lambda(2,1)[1 - \lambda(2,1) - z_1(1 - \lambda(2,2))]}{z_1[\lambda(2,2) - \lambda(2,1)]} \\ \lambda(1,2) = \dfrac{\lambda(2,2)[1 - \lambda(2,1) - z_1(1 - \lambda(2,2))]}{\lambda(2,2) - \lambda(2,1)} \\ \lambda(1,3) = \dfrac{[1 - \lambda(2,1) - z_1(1 - \lambda(2,2))]}{z_1(z_1 - 1)(\lambda(2,2) - \lambda(2,1))} \times \\ \qquad\qquad [(z_2 - 1)z_1\lambda(2,2) + \lambda(2,1)(z_1 - z_2)] \\ \lambda(2,3) = \dfrac{(z_2 - 1)z_1\lambda(2,2) + \lambda(2,1)(z_1 - z_2)}{z_2(z_1 - 1)} \\ \lambda(3,1) = \dfrac{\lambda(2,1)[z_3(1 - \lambda(2,1)) - z_1(1 - \lambda(2,2))]}{z_1(\lambda(2,2) - \lambda(2,1))} \\ \lambda(3,2) = \dfrac{\lambda(2,2)[z_3(1 - \lambda(2,1)) - z_1(1 - \lambda(2,2))]}{(\lambda(2,2) - \lambda(2,1))z_3} \\ \lambda(3,3) = \dfrac{[z_3(1 - \lambda(2,1)) - z_1(1 - \lambda(2,2))]}{z_1(z_1 - 1)z_4(\lambda(2,2) - \lambda(2,1))} \times \\ \qquad\qquad [(z_2 - 1)z_1\lambda(2,2) + \lambda(2,1)(z_1 - z_2)] \end{cases}$$

with constraints for $\lambda(2,1) < \lambda(2,2)$

$$(1 - \lambda(2,1))/(1 - \lambda(2,2)) > \begin{cases} z_1 > 1 \\ (1 - z_2)z_1/(z_1 - z_2) \\ z_1/z_3 > 1 \end{cases}$$

$$\lambda(2,1)/\lambda(2,2) < \begin{cases} z_1 < 1 \\ (1 - z_2)z_1/(z_1 - z_2) \\ z_1/z_3 < 1 \end{cases}$$

For $\lambda(2,1) > \lambda(2,2)$ a similar set of inequalities with opposite sign is obtained.